\PassOptionsToPackage{table}{xcolor}
\documentclass[dvipsnames,sigconf=false, nonacm=true, review=false, anonymous = false,]{acmart}
\usepackage{booktabs} 

\copyrightyear{2024}
\acmYear{2024}
\setcopyright{none}
\acmConference[GECCO '24]{2024
Genetic and Evolutionary Computation Conference}{July 10--14,
2024}{Melbourne, Australia}
\acmBooktitle{2024 Genetic and Evolutionary Computation Conference
 (GECCO '24), July 14--18, 2024, Melbourne, Australia}
\acmPrice{15.00}

\usepackage{amsmath}
\usepackage{xspace}
\usepackage{xcolor}
\usepackage{dsfont}
\usepackage{rotating}
\usepackage{makecell}
\usepackage{multirow}
\usepackage{subcaption}

\usepackage{amsthm}
\usepackage{amsmath}
\renewcommand{\epsilon}{\varepsilon}

\ccsdesc[500]{Theory of computation~Design and analysis of algorithms}
\ccsdesc[500]{Theory of computation~Bio-inspired optimization}

\begin{document}
\title[Large-scale Benchmarking]{Large-scale Benchmarking\\ of Metaphor-based Optimization Heuristics}



\author{Diederick Vermetten}
\affiliation{
  \institution{Leiden Institute for Advanced Computer Science}
  \city{Leiden}
  \country{The Netherlands}
}
\email{d.l.vermetten@liacs.leidenuniv.nl}

\author{Carola Doerr}
\orcid{0000-0002-4981-3227}
\affiliation{
  \institution{Sorbonne Universit\'e, CNRS, LIP6}
  \city{Paris}
  \country{France}}
\email{Carola.Doerr@lip6.fr}

\author{Hao Wang}
\affiliation{
  \institution{Leiden Institute for Advanced Computer Science}
  \city{Leiden}
  \country{The Netherlands}
}
\email{h.wang@liacs.leidenuniv.nl}

\author{Anna V. Kononova}
\affiliation{
  \institution{Leiden Institute for Advanced Computer Science}
  \city{Leiden}
  \country{The Netherlands}
}
\email{a.kononova@liacs.leidenuniv.nl}

\author{Thomas B{\"a}ck}
\affiliation{
  \institution{Leiden Institute for Advanced Computer Science}
  \city{Leiden}
  \country{The Netherlands}
}
\email{t.h.w.baeck@liacs.leidenuniv.nl}

\renewcommand{\shortauthors}{D. Vermetten et al.}

\begin{abstract}

The number of proposed iterative optimization heuristics is growing steadily, and with this growth,  there have been many points of discussion within the wider community. One particular criticism that is raised towards many new algorithms is their focus on metaphors used to present the method, rather than emphasizing their potential algorithmic contributions. Several studies into popular metaphor-based algorithms have highlighted these problems, even showcasing algorithms that are functionally equivalent to older existing methods. Unfortunately, this detailed approach is not scalable to the whole set of metaphor-based algorithms. Because of this, we investigate ways in which benchmarking can shed light on these algorithms.  To this end, we run a set of $294$ algorithm implementations on the BBOB function suite. We investigate how the choice of the budget, the performance measure, or other aspects of experimental design impact the comparison of these algorithms. 
Our results emphasize why benchmarking is a key step in expanding our understanding of the algorithm space, and what challenges still need to be overcome to fully gauge the potential improvements to the state-of-the-art hiding behind the metaphors. 

\end{abstract}

\maketitle

\section{Introduction}

When faced with an optimization problem, we have an ever-growing pool of algorithms to choose from. New optimization algorithms are continuously being proposed, which means that the challenge of understanding the state-of-the-art becomes harder by the day. For new approaches to stand out from everything else, researchers often refer back to metaphors as a framing device for their algorithms. 

The usage of metaphors to illustrate algorithmic ideas has been around for a long time and some of the most well-established algorithm families in the field made use of metaphors such as Darwin's theory of evolution in Evolutionary Computation~\cite{back1997handbook},
the swarming behavior of bird-like objects in the Particle Swarm Optimization algorithm~\cite{PSO} or the foraging behavior of ants in Ant Colony Optimization~\cite{DorigoPhDthesisACO}. 
Given the successes of these methods, it is natural that many new algorithms follow the same approach. There is, however, an increasingly visible problem, where the metaphor seems to become more important than the algorithm, which hinders the understanding of an algorithm's contribution to the state of the art~\cite{Sorensen15}.
In some cases, this leads to duplicated algorithms with different names, which grow to be highly cited and viewed as independent algorithms by practitioners~\cite{camacho2022analysis, StuetzleGreyWolf}. 

Even though many optimization algorithms are presented with an emphasis on the metaphor, there might still be interesting ideas and insights to be gained from understanding these algorithms in more detail. 
Given the widespread usage of these types of algorithms, writing off a method because of how it is presented seems counterproductive.

In this paper, we focus on benchmarking publicly available implementations  
of a wide variety of optimization algorithms. 
We specifically do not address the question of whether these algorithms contain any novelty in terms of algorithmic operators and only focus on showcasing their performance within a large benchmarking scenario. 
We highlight the ways in which benchmarking helps to gain insight into the strengths and weaknesses of optimization algorithms and discuss the potential benefits to be gained from such benchmark setups. In particular, we show that performance between algorithms is highly varied, with some algorithms performing consistently worse than RandomSearch, while others manage to outperform several well-established baselines. By considering two types of performance measures, we highlight the dependence of results on the used benchmarking setup. 
Finally, we discuss some of the challenges inherent to benchmarking studies, especially when performed on newly proposed algorithms. This includes questions regarding precise algorithm implementation, which causes seemingly the same algorithm to show widely different performance in our benchmark.

\section{Related Work}

Within the optimization field, metaphor-based optimization algorithms have been receiving quite some criticism in the last decade \cite{Velasco23MetaheuristicsReview}. 
One of the key arguments is that the usage of metaphors throughout an algorithm description does not advance our understanding of the algorithm, but only hides its true design ideas~\cite{Sorensen15, SorensenSG18}. To gain some idea of the scope of these types of algorithms, a community effort has been made in the form of the evolutionary computation bestiary, which documents the rise of metaphor-based algorithms over time~\cite{campelo2023lessons}. 

While detailed analyses of these optimization algorithms are time-consuming, in the last years several highly visible 
algorithms have been shown to contain no novelty over the previously existing algorithm families~\cite{StuetzleGreyWolf, Camacho-Villalon19, camacho2022analysis}. This has led to the community asking for stricter requirements when new algorithms are proposed~\cite{AranhaCCDRSSS22}, which are being adopted slowly, such as by the Transactions on Evolutionary Learning and Optimization (TELO) and Evolutionary Computation Journal (ECJ) journals where metaphor-based algorithms are now highly discouraged. 
On a wider scale, some initial benchmark studies have been performed recently, which highlight that many implementations of these metaphor-based algorithms perform very similarly to each other~\cite {del2021more, ma2023performance}. 
Finally, a study into some behavioral properties of these algorithms has shown that many of them are highly biased towards the center of the domain, leading to misleading performance comparisons if benchmarked on functions with similar types of bias~\cite{kudela2023evolutionary}. 

\section{Experimental Setup}

While nature-inspired optimization heuristics are common, it is often challenging to find open-source implementations of these algorithms which have been validated by the authors. 
We thus rely on third parties who have designed libraries containing a variety of algorithm implementations in relatively accessible formats. 
For this study, we identified four such libraries implemented in Python and used the following number of algorithms from them: 

\begin{itemize}
    \item 14 algorithms from EvoloPy~\cite{faris2016evolopy}, accessible at \url{https://github.com/7ossam81/EvoloPy}.
    \item 53 algorithms from Niapy~\cite{vrbanvcivc2018niapy}, accessible at \url{https://github.com/NiaOrg/NiaPy}.
    \item 139 algorithms from Mealpy~\cite{van2023mealpy}, accessible at \url{https://github.com/thieu1995/mealpy}.
    \item 76 algorithms from Opytimizer~\cite{de2019opytimizer}, accessible at \url{https://github.com/gugarosa/opytimizer}.
\end{itemize}

In addition to these libraries, we include some established algorithms as baselines, taken from the Nevergrad toolbox~\cite{nevergrad}. From Nevergrad, we utilize 8 algorithms: DE, diagonal CMA-ES, multiBFGS, 1+1 ES, PSO, Powell, Cobyla, and RandomSearch. Finally, we include two configurations from modular algorithm families: CMA-ES and BIPOP-CMA-ES from the modCMA package~\cite{modCMAGECCO} and DE and L-SHADE from the modDE package~\cite{modde}. To ease the analysis, we group these last 12 algorithms under the 'Baselines' denominator. All algorithms are used with default parameters settings from their respective libraries. 

In total, our portfolio consists of 294 algorithm implementations. Each of these algorithms is benchmarked on the single-objective, noiseless BBOB suite~\cite{bbobfunctions}, using the IOHexperimenter package~\cite{iohexperimenter}. For these experiments, we use all 24 functions contained in the BBOB suite, in dimensionalities $d\in\{2, 5, 10, 20\}$. For each function, we perform 5 independent runs on each of the first 10 instances. In total, this gives us $1\,411\,200$ runs. The budget for each of these runs is set to $B=10\,000\cdot d$.

Throughout this paper, we look at two different performance measures. The first is a standard fixed-budget setting, where we look at the precision (difference between best-so-far and optimal function values) at specific budgets. Since convergence curves are usually logarithmic, when aggregating these precisions we use geometric means unless stated otherwise. 
The second performance measure we use is an anytime measure: the normalized area over the convergence curve (AOCC), which is defined as follows: 
\begin{equation*}
    \textit{AOCC}(\Vec{y}) = \frac{1}{B} \sum_{i=1}^{B} \left( 1-\frac{ \min(\max((y_i), \textit{lb}), \textit{ub})  - \textit{lb}}{\textit{ub} - \textit{lb}} \right)
\end{equation*}
where $\Vec{y}$ is the vector of best-so-far precision reached during the optimization run, $B=10\,000\cdot d$ is the budget, $\textit{lb}$ and $\textit{ub}$ are the lower and upper bound of the precision values we consider. 
To be consistent with existing benchmarking studies on BBOB, when we refer to AOCC we make use of precision bounds $\textit{lb}=10^{-8}$ and $\textit{ub}=10^{2}$ with a logarithmic scaling between them. Since for higher dimensionalities $\textit{ub}=10^2$ can be a challenging bound on many functions, we also include some results on \textit{AOCC} with a relaxed upper bound of $\textit{ub}=10^8$, which we refer to as \textit{AOCClarge}. For all figures relating to \textit{AOCC} or \textit{AOCClarge}, an equivalent figure with the other bounds is available on our Figshare repository~\cite{reproducibility_and_figures}.

\subsubsection*{Reproducibility} To ensure the reproducibility of our work, our full benchmarking setup, raw and processed data and all scripts used for analysis and visualization presented in this paper are made available on Zenodo~\cite{reproducibility_and_figures}.

\section{Anytime Performance Results}\label{sec:anytime}

\begin{figure}
    \centering
    \includegraphics[width=0.5\textwidth]{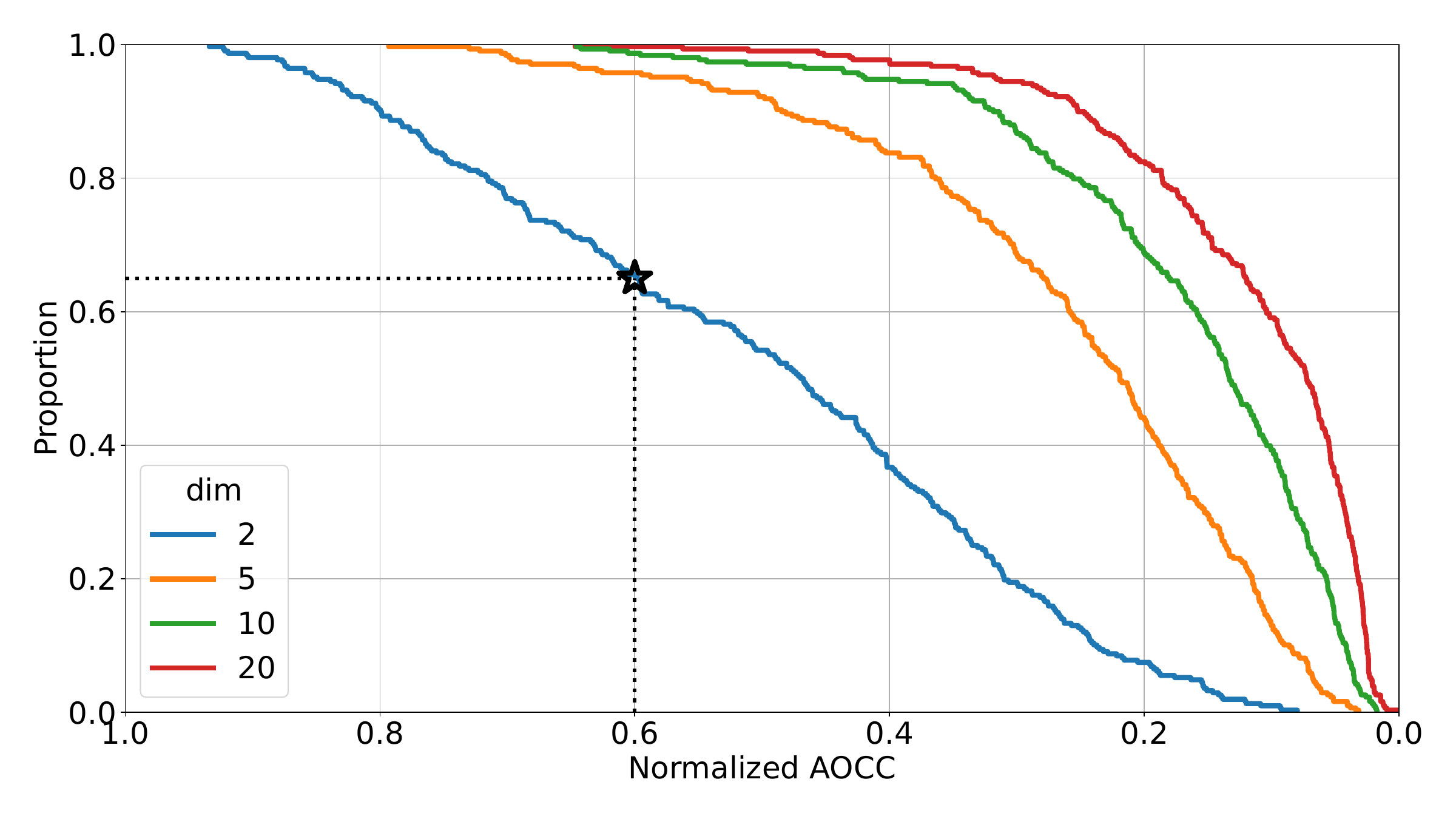}
    \caption{Cumulative Distribution of AOCC (default bounds) worse than $x$-axis value for all 294 algorithms in the portfolio. For example, in dimension 2, the star indicates that the fraction of algorithms with an AOCC below 0.6 is 0.65. AOCC values shown are aggregated over all 24 BBOB functions. }
    \label{fig:cumulative3}
\end{figure}

To investigate the overall performance of the selected algorithm portfolio, we look at the distribution of average AOCC across all functions, separated by problem dimensionality, which is visualized in Figure~\ref{fig:cumulative3}. From this figure, we can see that there is a rather wide spread of performance within the portfolio, especially for the lower dimensionalities. As dimensionality increases, not only do the functions become more challenging to optimize to the same precision, but the shape of the performance distribution changes as well. With growing dimensionality, there are some well-performing algorithms, after which average performance seems to drop off exponentially. This suggests that relatively few algorithms can scale effectively with regard to dimensionality. 

\begin{figure*}
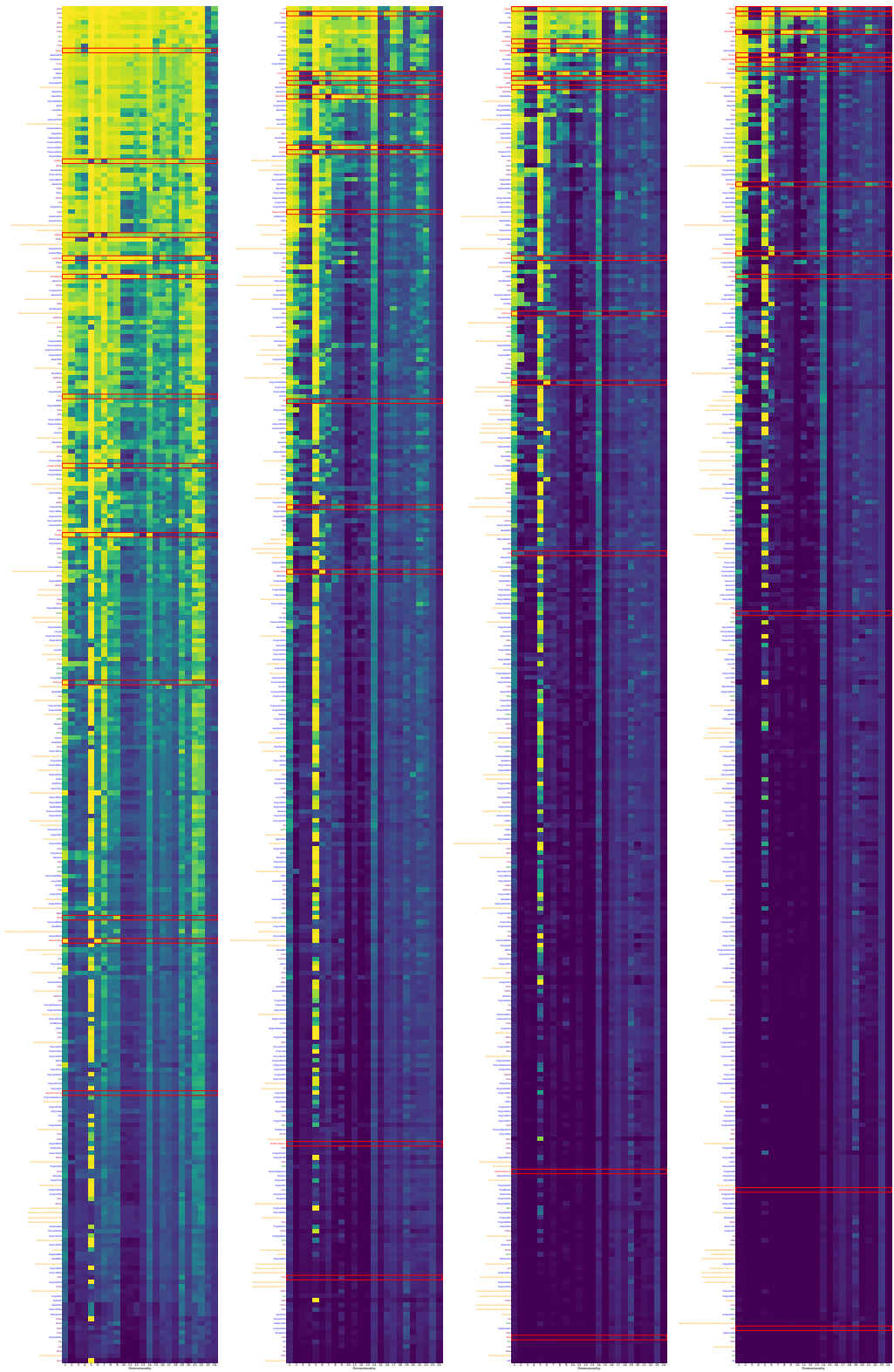

    \centering
    \includegraphics[height=0.95\textheight]{Figures/AOCC_2D_overview_SortFalse.pdf}
    \includegraphics[height=0.95\textheight]{Figures/AOCC_5D_overview_SortFalse.pdf}
    \includegraphics[height=0.95\textheight]{Figures/AOCC_10D_overview_SortFalse.pdf}
    \includegraphics[height=0.95\textheight]{Figures/AOCC_20D_overview_SortFalse.pdf}
    \caption{Normalized AOCC values per function for all 294 algorithms, ordered from dimensionality 2 (left) to 5, 10 and 20 (right). Color scales from dark blue=0 (worst) to yellow=1 (best). Larger versions of these figures are available on our Figshare repository~\cite{reproducibility_and_figures}. Colors denote the algorithm's library and algorithms are sorted by total AOCC over all functions.}
    \label{fig:AOCC results}
\end{figure*}

Since Figure~\ref{fig:cumulative3} can provide only a very highly aggregated view of the underlying performance data, we next look at the distribution of AOCC on a per-function and per-algorithm level. For each dimensionality, we create a heatmap indicating this per-function AOCC, shown in Figure~\ref{fig:AOCC results}. Since our portfolio consists of 294 algorithms, we highlight the baseline algorithm with a red rectangle. In addition, we color-code the libraries as follows: \textcolor{red}{Baselines}, \textcolor{purple}{Opytimizer}, \textcolor{orange}{Niapy}, \textcolor{green}{Evolopy} and \textcolor{blue}{Mealpy}. This coloring will remain consistent throughout all further figures. 
\begin{figure}
    \centering
    \begin{subfigure}{0.5\textwidth}
        \includegraphics[width=0.98\textwidth]{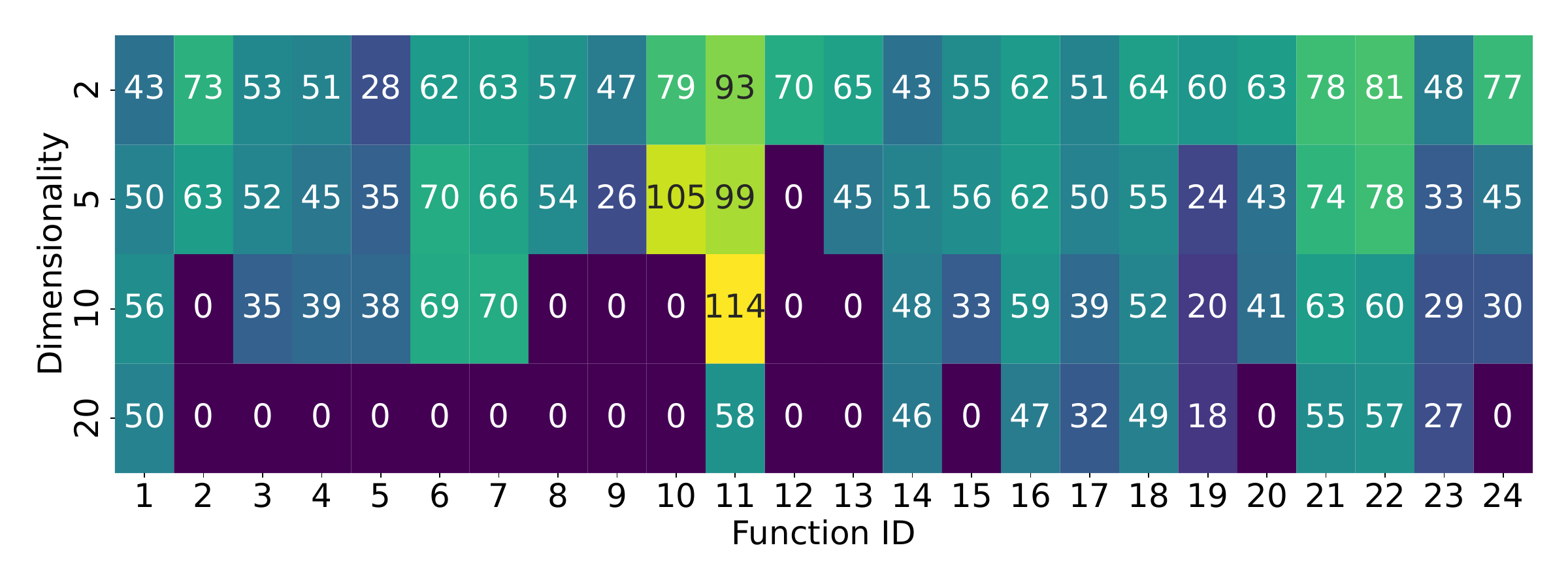}
        \caption{Number of algorithms (out of 294) which are worse than RandomSearch in each (function, dimensionality) combination.}
        \label{fig:rs_per_func}
    \end{subfigure}
    \begin{subfigure}{0.5\textwidth}
        \includegraphics[width=0.98\textwidth]{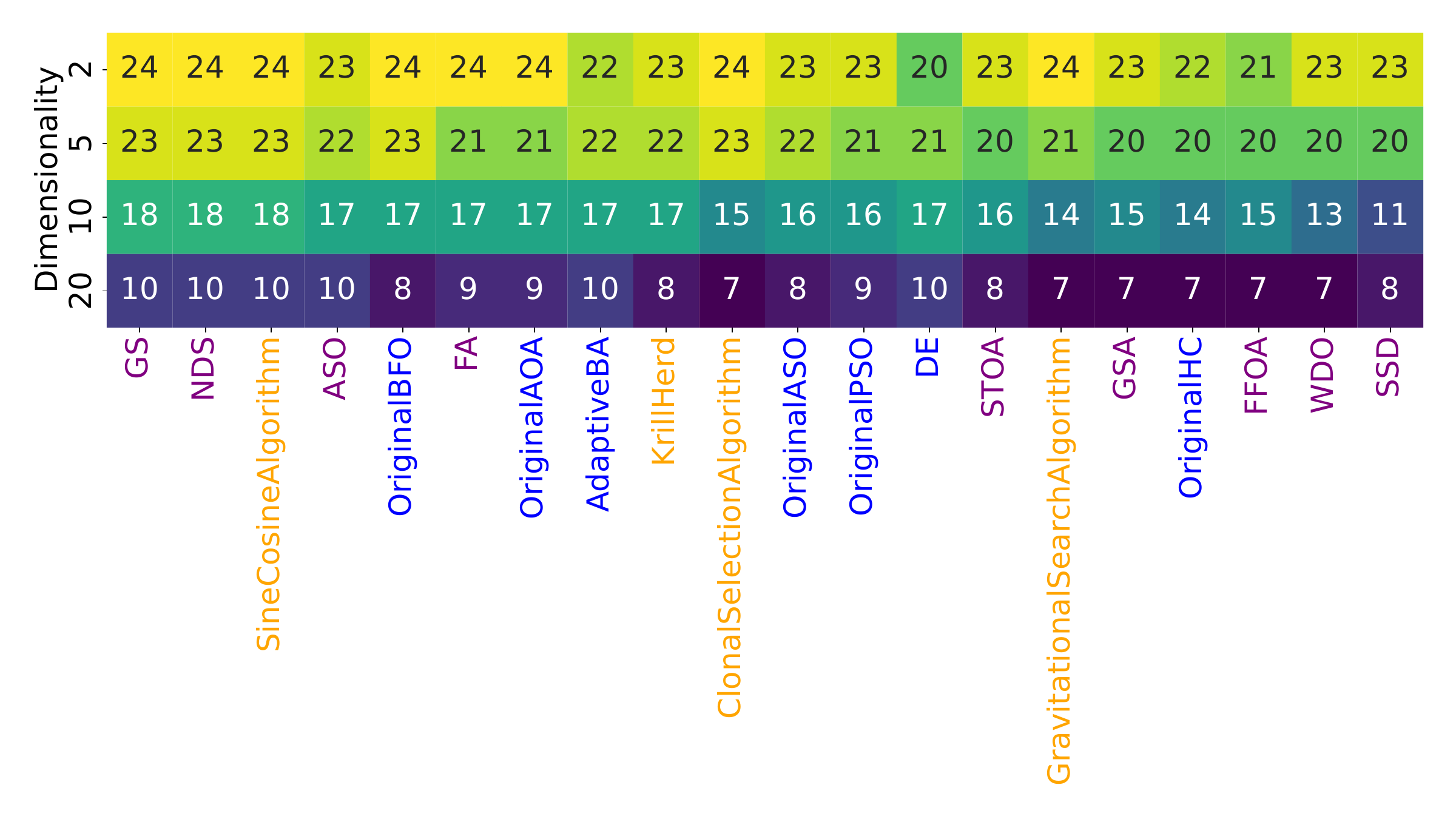}
        \caption{On how many functions (out of 24) the selected algorithms perform worse than RandomSearch in each dimensionality. Algorithms are selected based on the total number of functions on which the algorithm is worse than RandomSearch, with the top 20 of these algorithms included.}
        \label{fig:rs_top20}
    \end{subfigure}
    \caption{Comparisons of algorithms performance to RandomSearch based on AOCC. An algorithm is considered worse on a function if its AOCC is at least 10\% less than that of RandomSearch.}
    \label{fig:rs_comp_fraction}
\end{figure}

From Figure~\ref{fig:AOCC results}, we observe that, while several baseline algorithms are near the top, the best-performing algorithms come from a combination of different libraries. In general, there is no clear ordering between the libraries in terms of performance. When zooming in on the best algorithms, we also notice some patterns between the BBOB functions. For example, in dimensionality 10, the top-performing algorithm (BIPOP-CMA-ES), achieves relatively poor anytime performance on functions 3 and 4, while the second-best algorithm (JADE) manages those functions rather well. These kinds of differences highlight the potential complementarity between the algorithms in our portfolio. 

When looking at the worst performing algorithms in Figure~\ref{fig:AOCC results}, we note that RandomSearch is not the worst-performing algorithm. While the total number of algorithms which are worse on average decreases as the dimensionality grows, the total number of algorithms which fail to beat this baseline is not insignificant. To further analyze this aspect of our portfolio's performance, we identify per function how many algorithms achieve worse AOCC than RandomSearch by at least 10 percent, and show the results in Figure~\ref{fig:rs_per_func}. Since the ability of RandomSearch to hit the upper bound of AOCC decreases as the problem dimensionality grows, there are some functions where it has an AOCC of $0$, leading to no algorithms being considered worse. On some other functions, such as F11, it seems that RandomSearch is quite adequate in terms of anytime performance, beating over 30 percent of algorithms. For the remaining functions, there are a rather large portion of algorithms which compare poorly. Figure~\ref{fig:rs_top20} highlights the 20 algorithms which lose the comparison on the most functions. Given that some of these algorithms are worse on all 2-dimensional problems, it seems likely that their implementation is not fully functional. This confirms the importance of including RandomSearch as an algorithm to compare to in any benchmarking study. 
\begin{figure}
    \centering
    \includegraphics[width=0.5\textwidth]{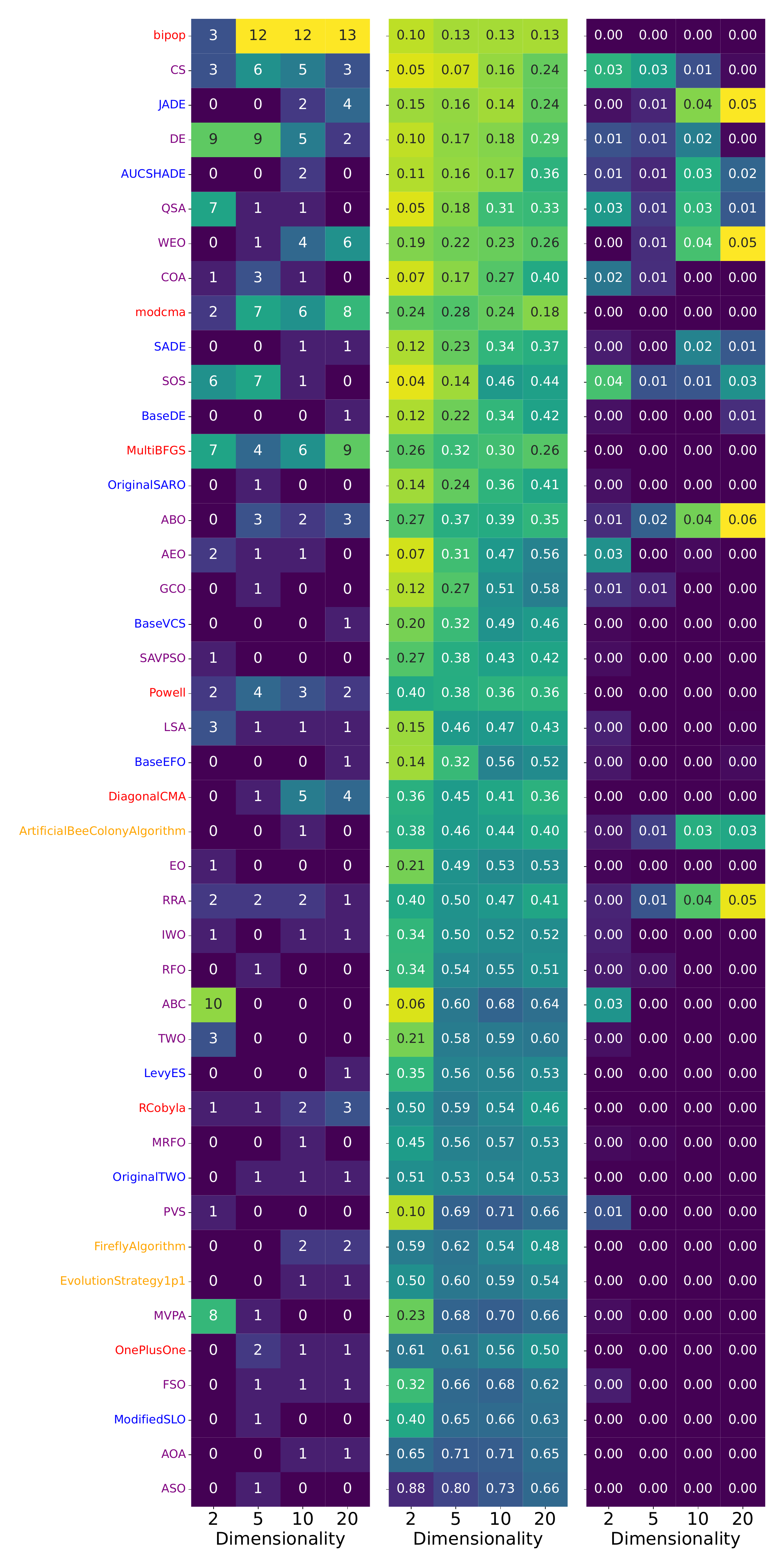}
    \caption{Left: Number of functions (out of 24) on which each algorithm is in the top 3 (on average AOCC). Middle: Average loss (absolute difference to best AOCC per function) over all 24 BBOB functions for each algorithm. Right: Contribution to an algorithm portfolio consisting of all baselines. Only algorithms which are considered competitive on at least one function are included, for a total of 43 unique algorithms. }
    \label{fig:top3_combined}
\end{figure}

Similar to the poorly performing algorithms, the left side of Figure~\ref{fig:top3_combined} zooms in on which algorithms achieve good performance on at least one function. Here, we characterize good performance as being ranked in the top 3 algorithms within our portfolio based on AOCC. While there are 96 (function, dimension) combinations, only 45 unique algorithms are in the top 3 for at least 1 function, and 20 of those show up exactly once. To gauge the overall performance of these algorithms, we calculate their total loss (difference in AOCC value to the best algorithm on each function, averaged over all functions), which is shown in the middle column of Figure~\ref{fig:top3_combined}.  While the top-performing algorithm in both figures is the same (BIPOP-CMA-ES), it is interesting to note the differences. Specifically, multiBFGS performs in the top for 26 (function, dimension) combinations, but in terms of average loss, it is only ranked 14th. This indicates that multiBFGS is a rather specialized algorithm, which leads to good performance on some types of problems, at the cost of worse performance on other function groups. On the other hand, the JADE algorithm is only in the top 3 for 6 (function, dimension) pairs, but ranks 3rd based on overall loss. Finally, we look at whether these algorithms could improve upon the set of baselines we consider. To achieve this, we consider a portfolio's performance to be the average of the minimum AOCC its component algorithms achieve on each function. By computing this measure for the set of all baselines, and for the set of the baselines with each considered algorithm included, we have a measure of contribution. These are shown on the right side of Figure~\ref{fig:top3_combined}, and highlight that purely considering the number of competitive functions ignores the scale of improvements, as can be seen for example in RRA, which is competitive in only 1 20-dimensional function, but contributes a lot to the 20-dimensional baseline portfolio. 

\section{Fixed-Budget Results}\label{sec:fb}

\begin{figure}
    \centering
    \includegraphics[width=0.45\textwidth]{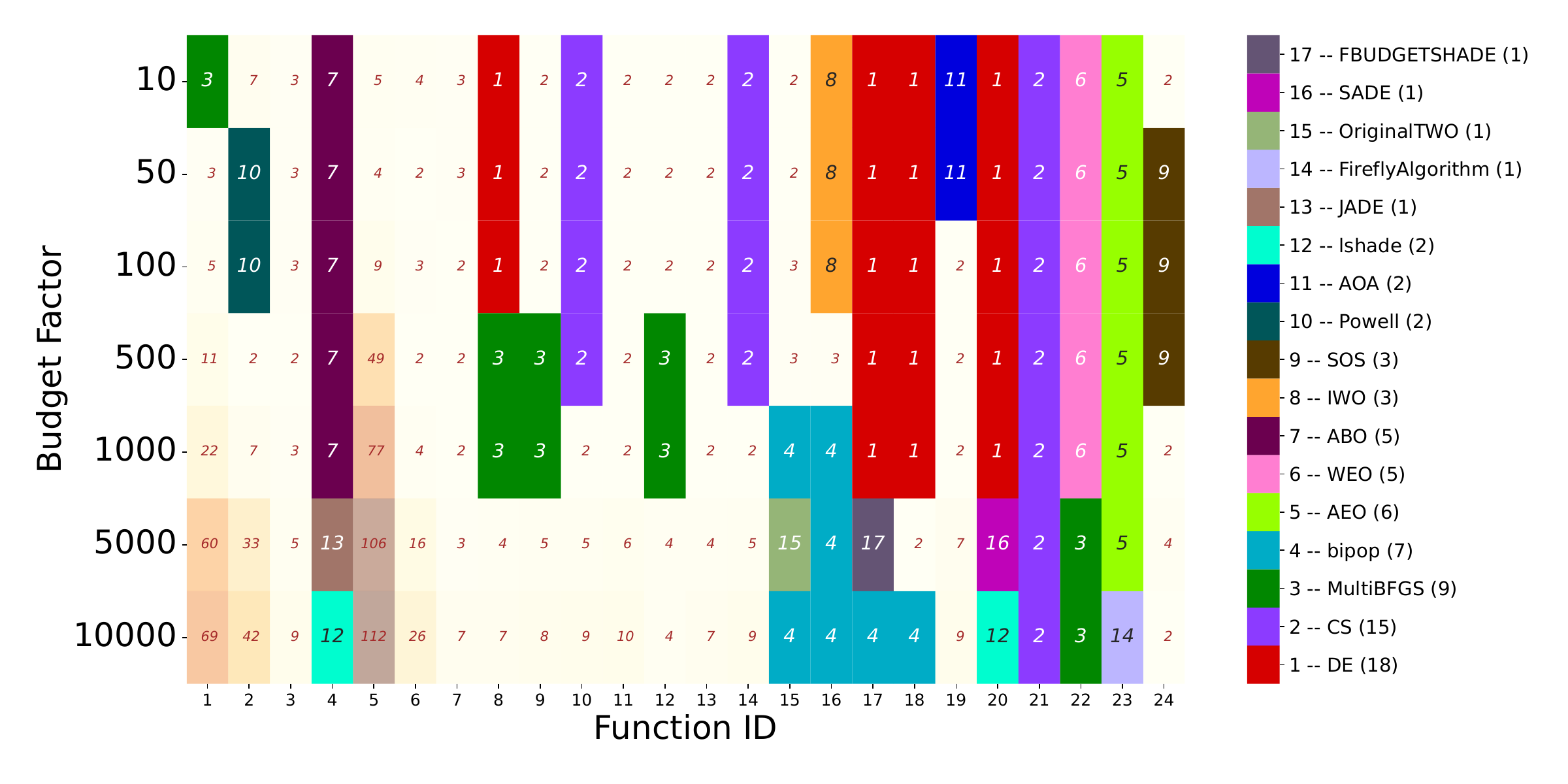}
    \caption{Best algorithm for each (budget, function) combination in dimensionality 10, based on average precision with a cutoff at $10^{-8}$. Light cells with brown text indicate a tie, with the number in the cell indicating how many algorithms are tied. The number in brackets after the algorithm name indicates how often it occurs in the figure.}
    \label{fig:best_budget4}
\end{figure}

In addition to the anytime performance metric, we can also analyze our data from a fixed-budget viewpoint. By recording the full performance trajectory during the benchmarking, we can compare different final budget values. This allows us to highlight the impact of the available optimization budget on the relative performance of the different optimization algorithms. For this analysis, we choose 7 different budget factors: $b\in\{10,50,100,500,1\,000,5\,000,10\,000\}$. For each dimensionality, the total budget is then set to $b\cdot d$. 

Since we observed in Figure~\ref{fig:top3_combined} that few algorithms were ever ranked in the top 3 on any given function, we start by analyzing the impact of budget on this finding. For a given dimensionality, we consider which algorithm is ranked first based on the average function value reached after the given budget has been used. This is visualized in Figure~\ref{fig:best_budget4}. Since we consider the problem optimized when a precision of $10^{-8}$ is achieved, ties can occur, especially on the `easier' problems with large budgets. As such, whenever 2 or more algorithms are tied, we don't make a distinction between them and instead only show how many algorithms are tied for that particular setting (shown in light cells with brown text). 

From Figure~\ref{fig:best_budget4}, it is clear that there are indeed many ties occurring, in particular for the sphere (F1) and linear slope (F5), where a large fraction of algorithms manages to reach the same function value after $10\,000\cdot d$ evaluations. We also observe some interesting patterns in terms of which algorithm ranks first on particular functions. For example, in F17 and F18 at low budgets, Opytimizer's DE 
performs well, but it is overtaken by BIPOP as the number of function evaluations increases. 

To better gauge how much each algorithm contributes to the overall performance of our portfolio, we make use of a Shapley-based analysis. In particular, we consider a fixed budget-factor and problem dimensionality, and with this setting we consider the performance of a set of algorithms to be the sum over all functions of the minimal average precision reached by an algorithm in the set on that function. The marginal contribution of an algorithm to a set is thus the difference in total precision when this algorithm is included and when it is excluded from the given set. By averaging this marginal contribution over 250 
sets of sizes between 1 and 20 (for a total of $5\,000$ sets), keeping the sets consistent across algorithms, we obtain an approximate Shapley value indicating to what extent the given algorithm contributes to the overall portfolio.

\begin{figure*}
    \centering
    \includegraphics[width=\textwidth]{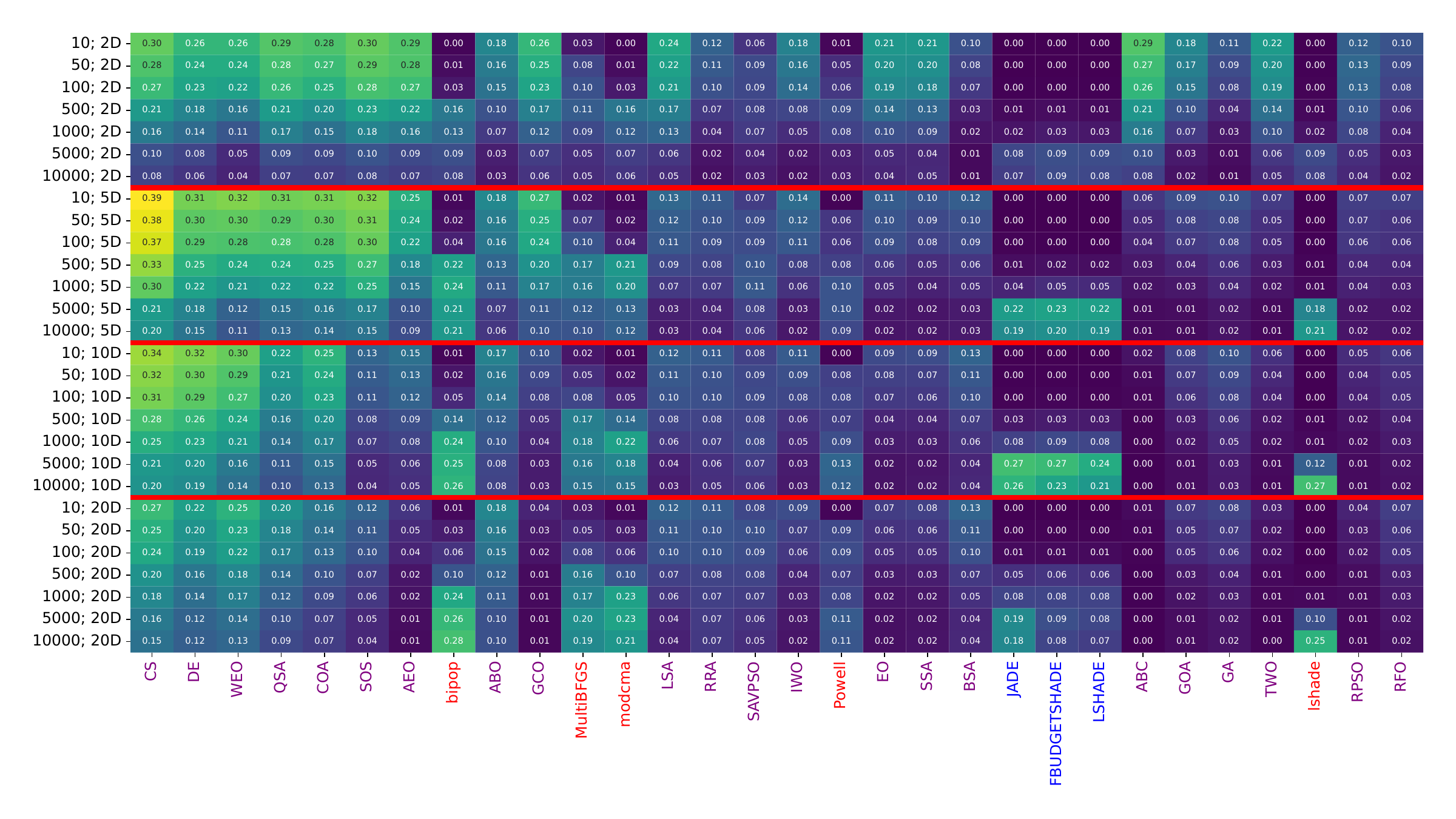}
    \caption{Normalized approximate Shapley values for each shown algorithm to portfolios of size at most 20, for different budget factors and dimensionalities. Algorithms are sorted based on total contribution across all functions, dimensions and budgets. Shapley values are computed based on fixed-budget contribution in log-space, capped at $10^{-8}$}
    \label{fig:shap_fbudget}
\end{figure*}

In Figure~\ref{fig:shap_fbudget}, we show the normalized version of these approximate Shapley values for 30 algorithms, for all dimensions and budget factors. These 30 algorithms were selected based on the best average approximate Shapley values over all settings. In this figure, we can observe a clear difference between algorithms which contribute more as budget increases (BIPOP, CMA-ES, several DE versions) and those which perform better at lower budgets, which are in the majority. When looking across dimensionalities, we see that, e.g., the ABC and GCO algorithms are rather effective in low dimensionalities, but fail to scale up effectively.

\section{Discussion} 

\subsubsection*{Understanding strengths and weaknesses of algorithms} As illustrated throughout this paper, algorithm performance data can be processed in a wide variety of ways. By varying the performance measure, type of comparison or level of aggregation, many different research questions can be investigated. While we are limited to only a few pages of results, providing access to the performance data allows others to re-use and expand on our analysis. For example, our data could be de-aggregated to only compare algorithm performance on a two-dimensional sphere to compare convergence speed between algorithms. The benefits of sharing performance data publically are clear to see when considering COCO's long-standing collection of performance data which comes from hundreds of algorithms, collected over more than a decade. Not only does this allow researchers to compare performance to a variety of known algorithms, but insights gained from looking at benchmark data can often inspire further research, and sometimes even lead to theoretical studies of empirically-found effects~\cite{DoerrYR0B18}.

A complementary aspect of understanding an algorithm's strengths is identifying whether it shows some performance characteristics which are not present in established algorithms. This might suggest that an algorithm contains useful new ideas or a way of combining existing ideas in a beneficial manner. This is particularly relevant for the type of nature-inspired optimization algorithms we discussed here, since many of them are introduced based on a metaphor, which obfuscates the underlying algorithmic concepts. In many cases, algorithms can be widely used  for years, to then be found to be equivalent to an existing algorithm with modified naming schemes. An example of this is the Cuckoo Search algorithm (CS), which performed especially well in this paper. However, it has been shown that CS contains no novelty, and is simply a reformulation of an existing ES variant~\cite{camacho2022analysis}. 

\begin{figure}
    \centering
    \includegraphics[width=0.5\textwidth]{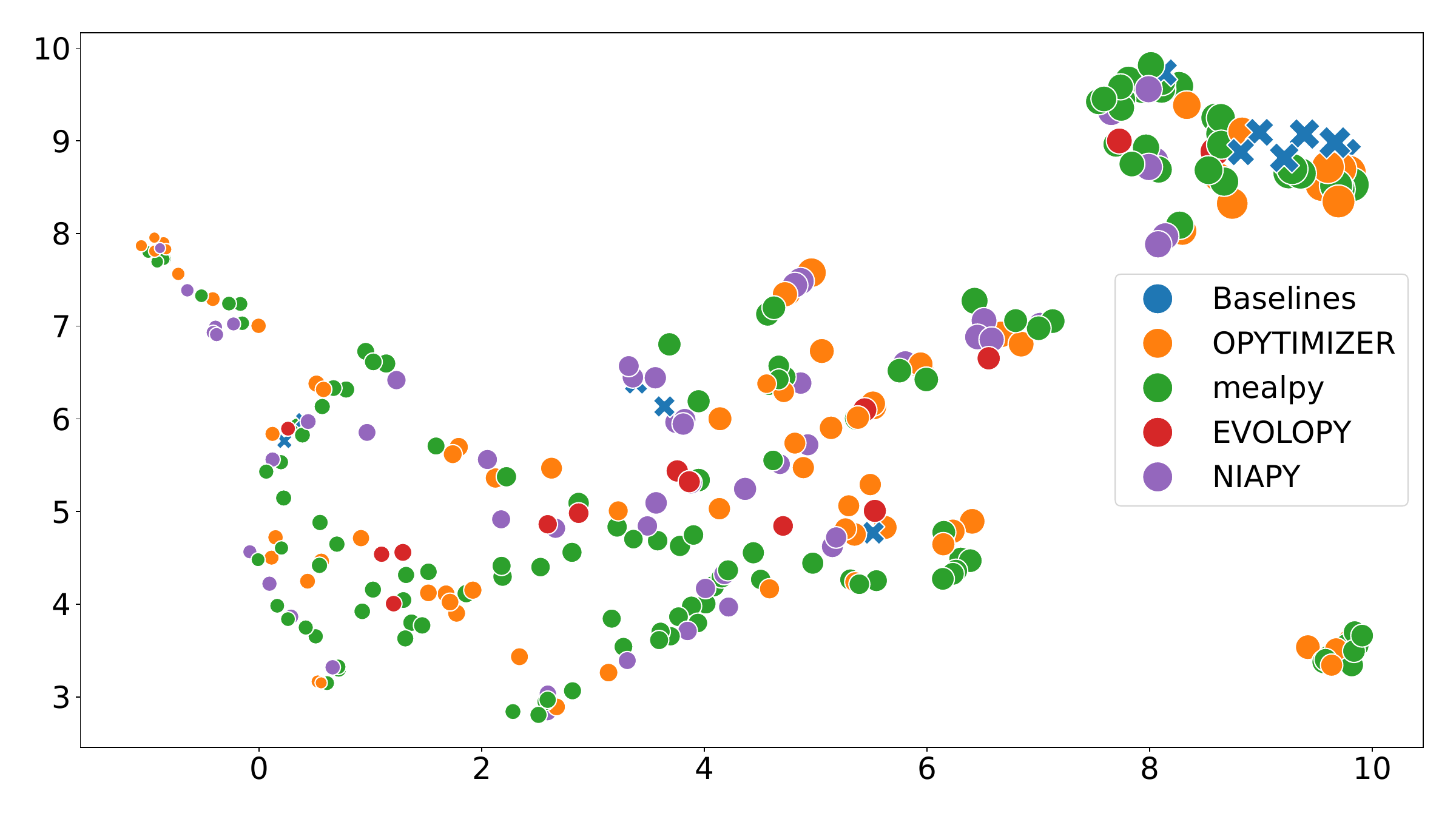}
    \caption{UMAP projection of the 96-dimensional (4 dimensionalities times 24 functions) AOCC vectors for each algorithm. Dots are sized based on average performance, with larger dots performing better. }
    \label{fig:umap_perf}
\end{figure}

\begin{figure}
    \centering
    \includegraphics[width=0.5\textwidth]{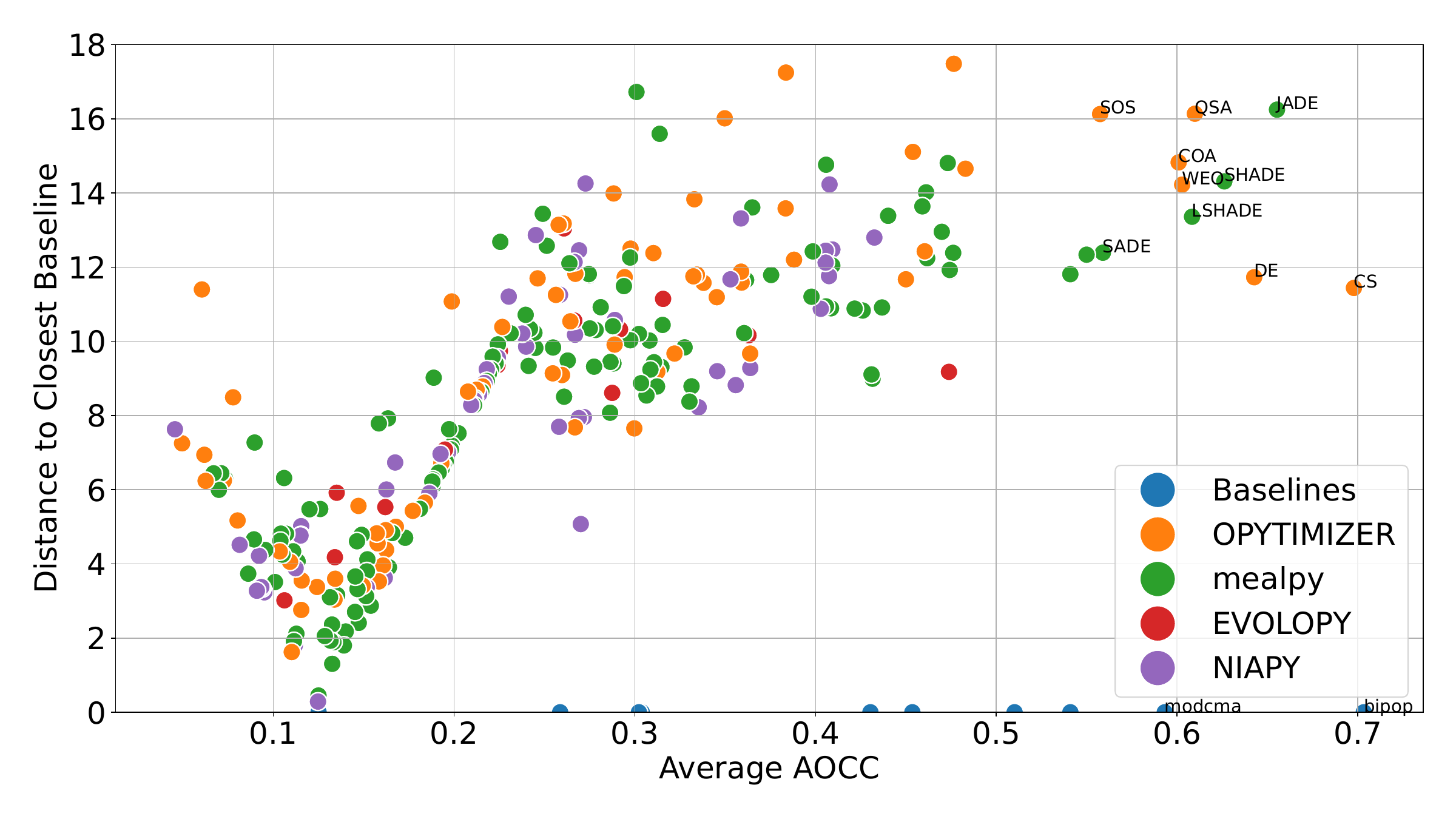}
    \caption{Relation between mean performance (AOCC averaged over all functions, dimensions) and distance to the closes baseline algorithm (manhattan distance). Algorithms with an average AOCC over $0.55$ are annotated.}
    \label{fig:dist_vs_perf}
\end{figure}

Since these detailed investigations into an algorithm's underlying principles can be time-consuming, benchmarking data offers a way to identify which algorithms might be worthwhile to analyze further. By looking for algorithm implementations which show strengths complementary to a set of baselines, we might discover the most high-potential algorithms from the larger portfolio. Given our benchmark data, we can look for algorithms which perform decently on average but are high-performing on a different set of (function, dimension) combinations than our baselines. To illustrate this approach, we can visualize the performance space using a dimensionality reduction technique, in our case UMAP~\cite{umap}, to reduce the performance representation of the algorithm down to 2 dimensions. In this case, the performance vectors are created by concatenating the per-function AOCC on all four dimensionalities, resulting in a 96-dimensional vector space. In the reduced version of this space, shown in Figure~\ref{fig:umap_perf}, we can see a cluster of well-performing algorithms on the top-right, which include several of the baseline algorithms. To zoom in on the relation between baselines and non-baselines, we can explicitly plot the average performance relative to the distance in performance space to the nearest baseline, as is done in Figure~\ref{fig:dist_vs_perf}.
Here, the algorithms in the top-right are of interest, since they both achieve good average performance as well as distinct performance differences from the closest baseline. These algorithms seem to be the most complementary to our baselines and could thus be the first candidates for further analysis. 

\subsubsection*{Data-driven algorithm selection} 
Since algorithms generally have different strengths and weaknesses, it makes sense to use this knowledge to make informed decisions on which algorithm to use to solve a given problem. Based on benchmark data, we can determine which algorithm performs best for given problem settings, where ``\emph{problem settings}'' can differ in resources available to solve the problems (budget, possibility to execute evaluations in parallel, etc) and on characteristics of the problems. This can result in recommendation systems based on aggregate performance over a varied set of functions, based only on problem dimensionality, budget, etc. A recent example of this is the NGopt wizard~\cite{NGOptMeunierTEVC22}. Alternatively, when considering the problem characteristics in more detail, for example by using part of the budget to compute low-level landscape features~\cite{mersmann2011exploratory}, we can exploit complementarity in performance on the function-level, or even the instance-level, to achieve automated algorithms selectors~\cite{KerschkeHNT19}.

\subsubsection*{Chaining rules} In addition to problem-level algorithm complementarity which can be exploited with algorithm selection methods, we can also use benchmark data to identify per-function complementarity. If one algorithm performs well at the beginning of the search but fails to converge to the optimum, while another algorithm takes a long time to find the right region but once found converges quickly, a hybrid algorithm which chains these two together could lead to significant improvements in anytime performance~\cite{VermettenBBOB,VermettenCMAdynAS}.

\subsubsection*{Judging Algorithmic Contributions}

Given the many different layers of algorithm complementarity which can be exploited using various learning methodologies, we should carefully consider how the contribution of an algorithm to the state of the art is judged. There is a clear distinction between generalist algorithms, which perform well across many functions, and specialist algorithms, which exploit specific function properties to achieve great performance on some, while losing out on others. Many newly proposed algorithms are studied from the generalist viewpoint, aiming to perform better than existing algorithms across wide suites of functions. This focus might mean missing out on interesting specialists, which could push the field forward by improving performance on a smaller set of functions. In communities like SAT-solving, competitions are now focussed more on contribution to a portfolio, rather than average performance, which has led to very potent solvers that select and combine the individual competitors~\cite{xu2008satzilla}. These approaches could be highly beneficial within optimization as well.

\subsubsection*{Underspecification of Parameterization / Implementations}

The algorithm portfolio we considered in this paper is a combination of algorithms from several different libraries, and as such some overlap between them is likely. We opted not to remove these duplicate algorithms, since the specific implementations might differ significantly. For example, many libraries contain some form of differential evolution (DE), which is a large algorithm family in its own right. An algorithm called DE could vary significantly based on which mutation operator or crossover variant is used, what adaptation rules and parameter settings are chosen or even how solutions outside the domain are handled~\cite{kononova2023importance}. Even for specific versions of DE, such as the L-SHADE algorithm~\cite{tanabe2014improving}, we observe differences between the implementation in modDE and Mealpy, as can be seen in Figure~\ref{fig:shap_fbudget}. In this case, the Mealpy version performs better at low budgets, while the modDE version becomes much more effective at larger budgets in higher dimensionalities. These differences once again highlight the need for code to be made available, as even detailed specifications like L-SHADE can result in these very different algorithm behaviors. 

When code is not available, and papers are the only source of an algorithm's specification, there are often insufficiently detailed or ambiguous descriptions of components, which make it almost impossible to fully recreate the used algorithm, resulting in difficulties in judging the accuracy of reported results. 
In addition to requiring reproducibility of results and availability of code, a robust way to judge algorithmic contributions could involve more focus on algorithm modularity. When proposing a new algorithm, it can often be framed as a modification of existing algorithms and thus implemented in existing modular frameworks, which are being proposed for many common algorithm families~\cite{modde, modCMAGECCO, camacho2021pso}. This way, the algorithm can be fairly compared to the base version of the used algorithm, as well as other modifications made available in the chosen framework, and the relative impact of different modifications can be analyzed rigorously~\cite{vanstein2024explainable}.

\subsubsection*{Algorithm Tunability}

While algorithm modularity can be a useful tool, it also raises another important question about the way in which parameter settings should be considered in comparative benchmarking studies. Most algorithms inherently contain a set of parameters which can drastically impact their performance on a chosen benchmark collection. This is especially obvious when comparing the settings achieved by hyperparameter optimization (HPO) on modular algorithms to their default values~\cite{modCMAGECCO}. If an algorithm is designed with tunability in mind, and its modules and/or parameters are set to perform well on a specific suite of functions, it should be no surprise if it outperforms a second algorithm tuned on a completely different set. As such, the question of which parameterizations were used should be kept in mind when drawing conclusions from benchmark data. In this study, we chose not to apply HPO to any of the considered algorithms, as the required computational effort would be too large. We believe that the data presented here could be used to guide more detailed follow-up work, which could address the question of how significant the results would change if HPO were to be applied equally to all algorithms.

\subsubsection*{Impact of Benchmarking Setup}

In Sections~\ref{sec:anytime} and \ref{sec:fb}, we looked at two different measures of algorithm performance. Many benchmarking studies, or papers in which new algorithms are introduced, tend to include only one type of analysis. As such, the choice of which performance measure to use is one of the many decisions which have to be made in order to present benchmark data. As can be seen when comparing, e.g., Figures~\ref{fig:top3_combined} and \ref{fig:shap_fbudget}, while some of the rankings between algorithms are consistent, they are by no means identical. Even within a specific performance perspective, other choices in the experimental setup, such as the available budget, all have an impact on the conclusions which can be drawn. One aspect in particular which is often cause for concern when a new algorithm is presented is the set of algorithms it is compared to. In the extreme case, one could present RandomSearch as a well-performing algorithm by comparing only to the implementations at the top of Figure~\ref{fig:rs_comp_fraction}, which would be misleading at best.  

While there is no right answer to the question of which performance measure, budget, function suite or baseline set is most appropriate for a given study, care should be taken to motivate these design choices, and their setting should be kept in mind when presenting conclusions about an algorithm's performance. 

\section{Conclusions and Future Work}

In this paper, we have illustrated the potential benefits that can be gained from robust benchmarking of optimization algorithms, in particular for the large set of metaphor-based optimizers. While many challenges remain to fully assess and fairly compare the contributions made by these algorithms, our analysis highlights that benchmarking can shed light on the relative strengths of algorithms, identifying the most interesting candidates for follow-up studies. We have also identified a surprisingly large number of algorithms that struggle to outperform random sampling on problems that can be considered very easy.

To reduce the burden of setting up a sound benchmarking environment, a number of open-source software tools are being developed to reduce the barrier of entry to rigorous benchmarking. With tools like COCO~\cite{hansen2021coco}, Nevergrad~\cite{nevergrad}, IOHprofiler~\cite{IOHanalyzer}, and many more, it is now easier than ever to run benchmarking studies and to compare performance and behavioral data to that of hundreds and even thousands of previously evaluated ones. The use of standardized benchmarking practices and data recording practices also facilitates reproducibility and data sharing~\cite{lopez2021reproducibility}.    

While our study focuses on the BBOB problem suite, it should be noted that we are not claiming that all studies should be run on this same benchmark suite. While there is some benefit in terms of comparability with existing data, sticking with a single benchmark suite might risk overfitting to the biases of that suite, which might result in worse generalizability. What matters to us is that algorithms are assessed in a fair manner, on problems that allow to assess strength and weaknesses of the algorithms in different optimization scenarios.  

\begin{acks}
    This work was supported by CNRS Sciences informatiques via the AAP project IOHprofiler. 
\end{acks}

\bibliographystyle{ACM-Reference-Format}
\bibliography{refs}

\end{document}